\pdfoutput=1

\documentclass[11pt]{article}

\usepackage[]{EMNLP2022}

\usepackage{times}
\usepackage{latexsym}
\usepackage{colortbl}

\usepackage[T1]{fontenc}

\usepackage[utf8]{inputenc}

\usepackage{microtype}

\usepackage{inconsolata}

\usepackage{multirow}
\usepackage{graphicx}
\usepackage{amsmath}
\usepackage{amssymb}
\usepackage{comment}

\newcommand*\samethanks[1][\value{footnote}]{\footnotemark[#1]}

\title{
RLET: A Reinforcement Learning Based Approach for Explainable QA with Entailment Trees}

  \author{Tengxiao Liu\textsuperscript{1}\thanks{{} {} Work done during internship at AWS Shanghai AI Lab.}, 
  Qipeng Guo\textsuperscript{2},
  Xiangkun Hu\textsuperscript{2}, 
  Yue Zhang\textsuperscript{3}\thanks{{} {} Corresponding authors.}, 
  Xipeng Qiu\textsuperscript{1}\samethanks, 
  Zheng Zhang\textsuperscript{2}\\
  \textsuperscript{1}School of Computer Science, Fudan University \\
  \textsuperscript{2}Amazon AWS AI,
  \textsuperscript{3}School of Engineering, Westlake University\\
  \texttt{txliu21@m.fudan.edu.cn, }
  \texttt{\{gqipeng, xiangkhu, zhaz\}@amazon.com} \\
  \texttt{xpqiu@fudan.edu.cn, } \texttt{zhangyue@westlake.edu.cn}\\}

\begin{document}
\maketitle
\begin{abstract}


Interpreting the reasoning process from questions to answers poses a challenge in approaching explainable QA. A recently proposed structured reasoning format, entailment tree, manages to offer explicit logical deductions with entailment steps in a tree structure. 
To generate entailment trees, prior single pass sequence-to-sequence models lack visible internal decision probability, while stepwise approaches are supervised with extracted single step data and cannot model the tree as a whole.
In this work, we propose RLET, a \underline{R}einforcement \underline{L}earning based \underline{E}ntailment \underline{T}ree generation framework, which is trained utilising the cumulative signals across the whole tree. 
RLET iteratively performs single step reasoning with \textit{sentence selection} and \textit{deduction generation} modules, from which the training signal is accumulated across the tree with elaborately designed aligned reward function that is consistent with the evaluation.
To the best of our knowledge, we are the first to introduce RL into the entailment tree generation task. Experiments on three settings of the EntailmentBank dataset demonstrate the strength of using RL framework.

\end{abstract}

\section{Introduction}

\begin{figure}[t]
    \centering
    \includegraphics[width=1.0\linewidth]{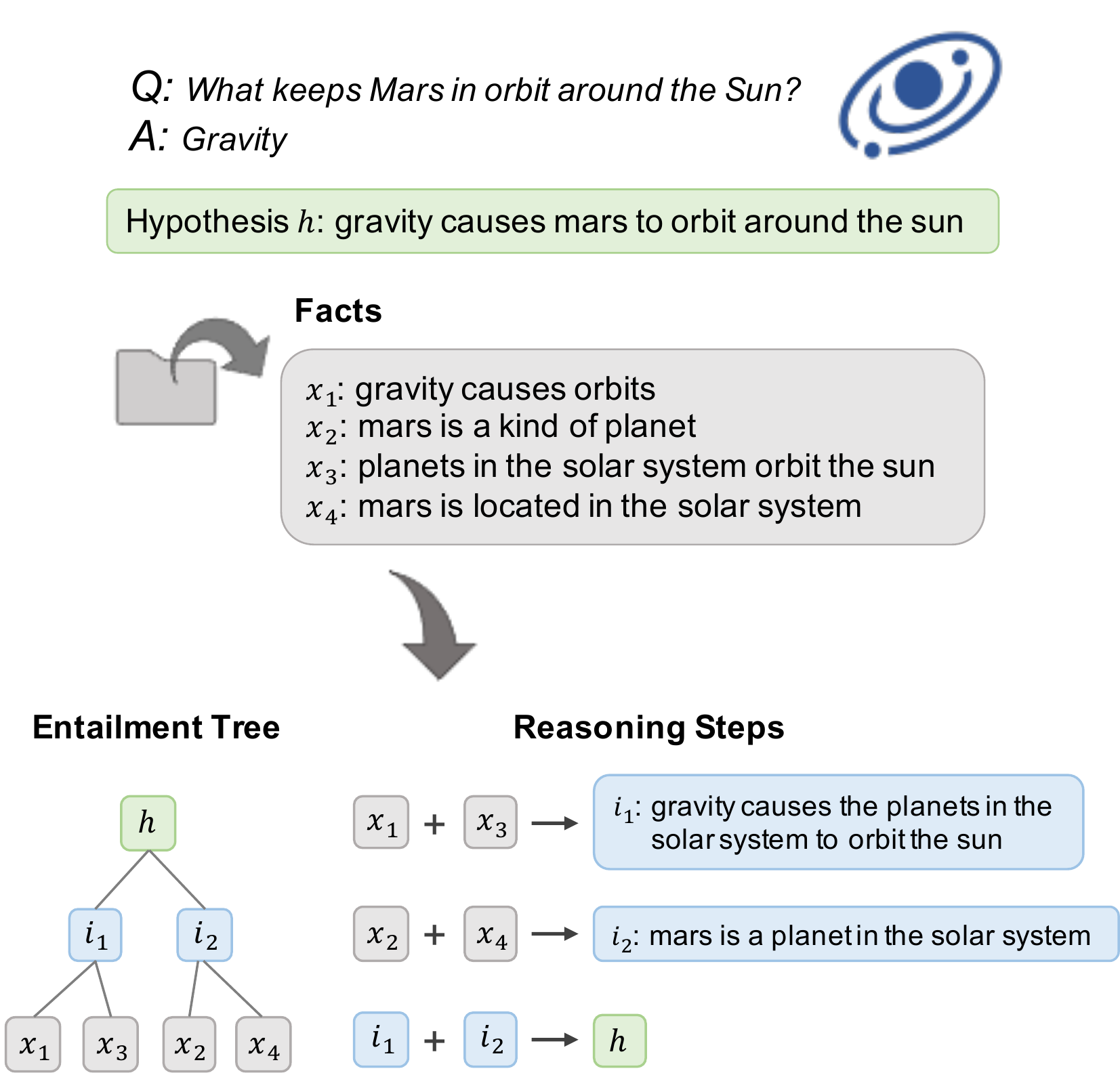}
    \caption{The entailment tree generation task. Given QA pair, hypothesis (green boxes) and fact sentences (grey boxes), the model needs to generate tree-structured reasoning chain along with natural language intermediate conclusions (blue boxes).}
    \label{fig1}
\end{figure}

Reasoning over explicitly given knowledge and generating detailed deduction steps are important challenges towards the goal of automated reasoning in AI community \citep{McCarthy1960ProgramsWC, Wos1985WhatIA, Mercier2011WhyDH}. 
Interpreting the reasoning process in QA systems can provide human-understandable inspections into the logical deductions and help ensure the soundness and reliability of the models \citep{Wiegreffe2021TeachMT}. 
As shown in Figure~\ref{fig1}, given the question ``What keeps Mars in orbit around the Sun?'', a reasoning process can explain why the answer ``Gravity'' is predicted, thereby strengthening the trustworthiness.

Recent work in the field of explainable QA includes extracting keywords or sentences that indicate the answers \citep{Yang2018HotpotQAAD}, constructing multi-hop explanations from knowledge graph \citep{Xu2021DynamicSG, Xie2020WorldTreeVA} and generating free-form explanations and reasoning chains \citep{Camburu2018eSNLINL, Rajani2019ExplainYL,Jhamtani2020LearningTE, Wei2022ChainOT, Wang2022SelfConsistencyIC}. 
Among various explanation approaches, multi-step entailment trees \citep{Dalvi2021ExplainingAW} build tree-structured reasoning chains and explicitly demonstrate the deduction steps from given knowledge to the hypothesis, which can offer interpretability in more detail (Figure~\ref{fig1}). 

One line of previous work considers the entailment trees as linearised sequences and adopt sequence-to-sequence (Seq2Seq) models to generate the entire reasoning chain in a single pass with all given sentences as input \citep{Dalvi2021ExplainingAW, Tafjord2021ProofWriterGI}. In these approaches, the internal steps lack trustworthiness because the probability distribution remains invisible and not controllable. 
Besides, each deduction step has access to the full input so it might contain information leaked across all input sentences instead of only deducing over its own premises. 
To alleviate these problems, subsequent work splits the tree into multiple steps and trains the model to perform single step reasoning \citep{Ribeiro2022EntailmentTE, Bostrom2022NaturalLD, Hong2022METGENAM}. These models, however, are trained with isolated single-step data, which ignores the dependencies between steps as a whole chain. Since new intermediate sentences will be generated at each step, every action poses an impact on its subsequent decisions. 
The absence of such consideration in training will 
lead to the inconsistency with evaluation which scores the chain in a comprehensive way. 


The reasoning chain is a sequence of discrete actions. As such,
we address the above issues by presenting RLET,  a  \underline{R}einforcement \underline{L}earning (RL) based \underline{E}ntailment \underline{T}ree generation framework that models the entire reasoning chain as a Markov Decision Process (MDP).
Specifically, we decompose the task into two parts: sentence selection and deduction generation. At each step, the model will first select two sentences (including both given facts and generated intermediate conclusions) for composition, and the deduction generation will combine them into a new intermediate conclusion and add it to the next step.
After constructing a whole chain, each step will receive a reward depending on its contribution to the overall correctness and validity of the entire tree.

Enjoying the convenience of crafting reward functions in RL, we can flexibly assign evaluation-consistent rewards to the steps, instead of purely relying on exact match with the gold tree.
In such a way, model behaviors can be manipulated with more flexibility, getting rid of the rigorous chronological match with ground truth.
Supervised by the cumulative rewards, the model is encouraged to find the optimal policy that leads to greater good. 
Such advantages can not only bring improvements on post-hoc explanation modeling, but also benefit interpretable model decision making, where the reasoning process is integrated into the inference process.
Extensive experiments on three settings of the benchmark dataset EntailmentBank \citep{Dalvi2021ExplainingAW} demonstrate that our approach  outperforms existing baseline models on structure correctness and verify the strength of embracing the end-to-end RL framework by modeling the entire reasoning chain.
To our knowledge, we are the first to introduce RL framework into the task of entailment tree generation, taking care of the entire chain with the aid of flexible reward design that
aligns with the final evaluation metric, bridging the gap between training objective and evaluation results. \footnote{Code is publicly available at: \url{https://github.com/tengxiaoliu/RLET}.}

\section{Method}

\begin{figure*}[t]
    \centering
    \includegraphics[width=1.0\linewidth]{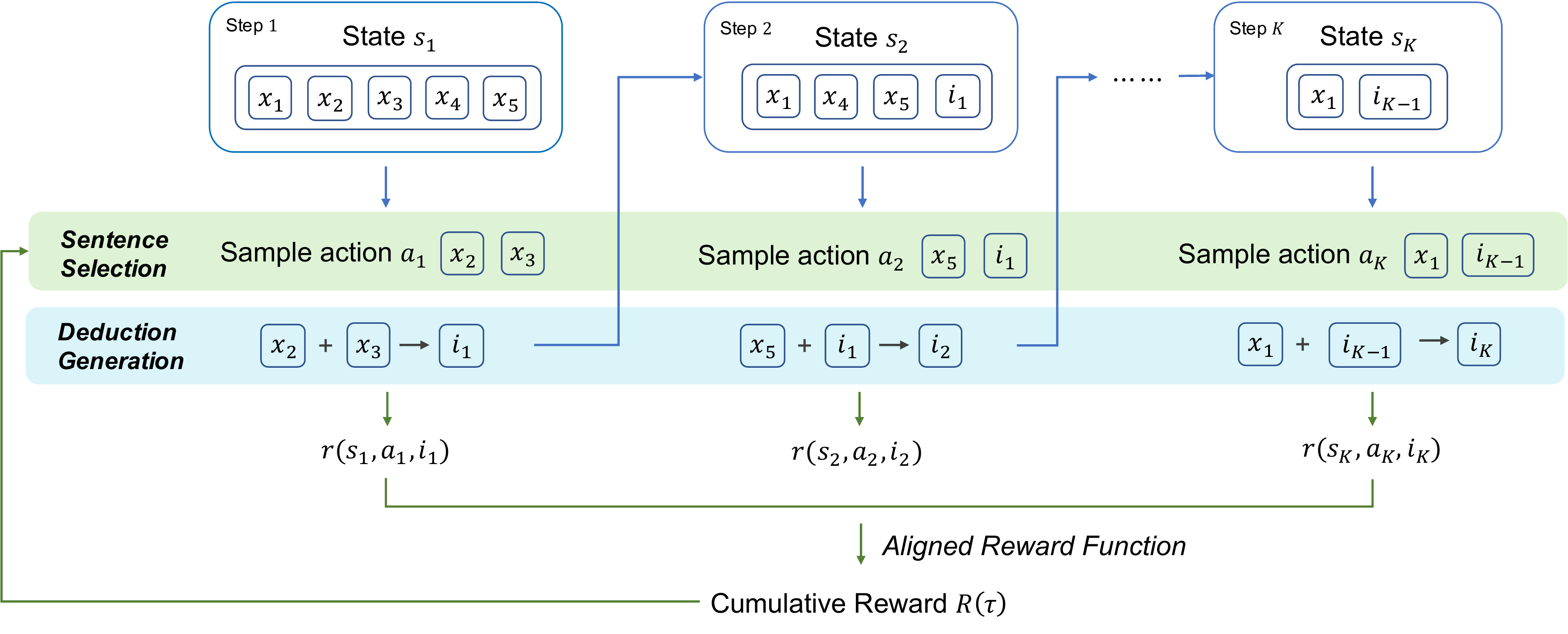}
    \vspace{-10pt}
    \caption{\label{fig_model} Reasoning pipeline of RLET. At each reasoning step, sentence selection module will select two sentences from its state for composition, and the deduction generation will yield the intermediate conclusion deducted from them. The next state will exclude the selected sentences and add the new intermediate conclusion. After generating the full trajectory, the cumulative reward will be used to update the parameters of sentence selection module.}
\end{figure*}

Our goal is to provide a step-by-step reasoning process for commonsense science questions, with prior knowledge of the question, the correct answer and a set of fact sentences. We first describe the task formulation and then explain each part in detail.

\subsection{Task Formulation}
We formulate the reasoning process as an entailment tree construction task with each step as a logical entailment.
The inputs of our task include a collection of fact sentences $X =\{x_1, x_2, \cdots, x_n\}$ and a hypothesis $h$, where $X$ consists of both relevant and non-relevant sentences.
Specifically, the hypothesis $h$ is the combination of the question and its correct answer, stated in a declarative form.
The task aims to construct an entailment  tree $T$ from the bottom up, with selected facts from $X$ as leaf nodes, the hypothesis $h$ as the root node and generated intermediate conclusions as internal nodes $I$. 
Each intermediate conclusion $i_k \in I$ is generated by deducing from its immediate children during the construction of the tree. A reasoning step includes selecting the premises and generating the intermediate conclusion.
Following \citet{Bostrom2022NaturalLD}, we only consider two-sentence combination at each reasoning step for simplicity. 
\footnote{Given the fact that any multi-way tree can be transformed into a binary tree, it is possible to apply this assumption.}
At each reasoning step, we further decompose it into two sub-tasks: sentence selection and deduction generation. 

\paragraph{Sentence Selection}
The goal of the sentence selection 
module is to choose two sentences 
as premises of single step reasoning.
At step $k$, sentence selection takes hypothesis $h$, fact set $X$ and intermediate set $I^k$ as input, where $I^k = \{ i_1, \dots, i_{k-1}\}$ denotes the intermediate conclusions generated in previous $k-1$ steps.
The desired output includes two sentences (nodes) $\{n_i, n_j\}\subseteq X \cup I^k$ that will make a logical combination in the following module.


\paragraph{Deduction Generation}
Given the selected sentences as input, deduction generation
outputs a new intermediate conclusion 
$i_k$ deduced from these two sentences
: $i_k = g(n_i, n_j)$, where $g$ denotes a Seq2Seq model. 
The conclusion should be well entailed by the premises, and reasons over the information from the given sentences only.






\subsection{RL-based Sentence Selection}

The design ethos of our approach is to bridge isolated single steps  with cumulative training signals, which fits very well with the nature of RL.
To tackle sentence selection task, we model the entailment tree as a Markov Decision Process, which can be denoted as 
a tuple $(\mathcal{S}, \mathcal{A}, R, T)$, where $\mathcal{S}$ is a set of states, $\mathcal{A}(s)$ is the action space at state $s$, $R(\cdot)$ is the reward function and $T(\cdot)$ represents the transition function. Our goal is to learn an optimal policy $\pi$ that decides what action to take at each state.

As shown in Figure~\ref{fig_model}, at reasoning step $k$, the \textbf{state} $s_k$ is defined as a collection of unused sentences (nodes) including both facts and generated intermediate conclusions, namely $s_k = \{ n \mid n \in X \cup I^k \setminus U^k \}$, where $U^k$ is the set of used sentences in previous steps. 
For example, at the very beginning $k = 1$, the initial state $s_1$ should contain all the given fact sentences $X$. 
Correspondingly, the \textbf{action space} $\mathcal{A}(s_k)$ at state $s_k$ is all the potential actions at the current step, where an action is a pair of sentences within the state. Formally, the action space is represented as
$\mathcal{A}(s_k)=  s_k \times s_k$.

With a pre-trained DeBERTa \citep{He2021DeBERTaV3ID} model as the backbone encoder of the sentence selection module, we encode each action candidate in the form of \texttt{[CLS]} $h$ \texttt{[SEP]} $a_k^i$ and apply a feed forward neural network together with Softmax activation on the \texttt{[CLS]} representation to obtain the normalised probability distribution over the action space, where $a_k^{i}$ is the concatenation of the two sentences.
Formally, the probability of each action $a_k^{i}$ can be described as:
\begin{equation}
    b^i_k = \text{FFN}(f([h,a^i_k])), 
\end{equation}
\begin{equation}
    \pi_\theta (a_k^{i} \mid s_k) = \text{Softmax}(\{b^i_k\}),
\end{equation}
\noindent where $\theta$ is the parameters of policy $\pi$, $f(\cdot)$ denotes the contextualised representation of the \texttt{[CLS]} token, $[\cdot]$ denotes concatenation, $b^i_k$ is the score of action $a_k^{i}$.
At step $k$, we then sample one action $a_k$ based on the probability distribution:
\begin{equation}
    a_k \sim \pi_{\theta}( \boldsymbol{a_k} \mid \mathcal{A}_k, s_k ).
\end{equation}

Given the two sentences within the sampled action, the deduction generator performs logical combination and outputs an intermediate conclusion ($i_1$ in Figure~\ref{fig_model}). Accordingly, the state $s_k$ will be shifted to $s_{k+1}$, with the selected sentences being removed, and the newborn sentence added. Meanwhile, the action space will also alter to include the action candidates composed from current state. In our task, the \textbf{state transition} is always deterministic with the transition function $T(s_k, a_k, s_{k+1}) = 1$.

In summary, we represent a step as $(s_k, a_k, i_k)$, meaning taking action $a_k$ at state $s_k$ and generating the intermediate conclusion $i_k$.
After undergoing several iterations of the sentence selection and deduction generation, we will obtain a trajectory denoting the reasoning steps we take to construct the entailment tree
\begin{equation}
    \tau = \{s_1, a_1, i_1, s_2, a_2, i_2, \dots, s_K, a_K, i_K\},
\end{equation}
where $K$ is the length of the reasoning chain. 

\paragraph{Reward}\label{sec_reward}
\begin{figure}[t]
    \centering
    \includegraphics[width=\linewidth]{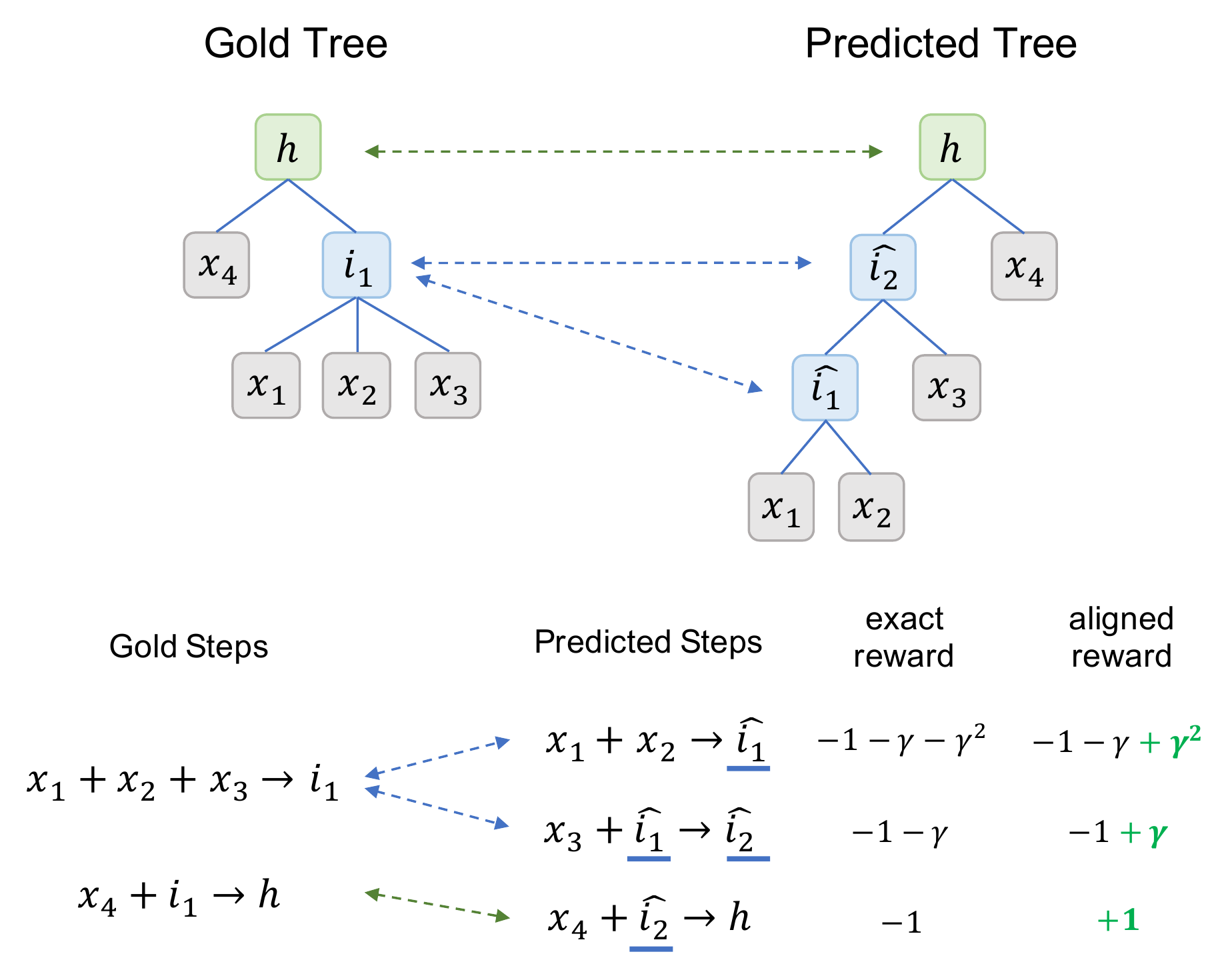}
    \vspace{-10pt}
    \caption{\label{fig_align} Illustration of applying alignment algorithm between a predicted tree and its gold tree. Dashed lines represent the connected intermediate nodes are aligned. Underlined nodes $\hat i_1, \hat i_2$ in predicted steps are both aligned to gold internal node $i_1$, which finally leads to a positive signal in the cumulative aligned reward. $\gamma$ denotes the discount factor. }
\end{figure}


Before evaluating entailment trees, an alignment algorithm \citep{Dalvi2021ExplainingAW} will be applied to guarantee the flexibility of the tree structure evaluation.
However, 
stepwise models in previous work are only supervised by the exact match with single-step ground truth in training, resulting in inconsistency with actual evaluation.


To address this issue, we involve the alignment algorithm in the design of our reward function.
To begin with, for each intermediate node in our trajectory and the gold tree, we gather all leaf nodes in their children respectively. We consider one generated internal conclusion (node) $\hat i$ is aligned to a gold internal node $i$ if the leaf nodes in their children have the maximum Jaccard similarity. If $\hat i$ has zero similarity with every gold internal node, we align it to a blank node with no conclusion. 
We assume the aligned nodes are similar in semantics since they are reasoned from similar facts.
As described in Figure~\ref{fig_align}, the predicted intermediate nodes are both aligned to the gold node $i_1$ according to their Jaccard similarity scores.

Rewards will be assigned to each reasoning step after the full trajectory is generated.
For each step $(s_k, a_k, i_k)$, we give it an independent reward $r_k$ based on the exact match between its action and the premises in the aligned gold step. 
\begin{equation}
  r_k=\begin{cases}
    1 , & \text{if perfect match}.\\
    -1 , & \text{otherwise}.
  \end{cases}
\end{equation}

Without the aid of the alignment, the predicted steps in Figure~\ref{fig_align} shall all receive negative exact rewards because their chosen premises fail to have an exact match with the gold steps. 
In contrast, after aligning, the last predicted step gets a positive reward once $\hat i_2$ is aligned with $i_1$. 


        

Then the final cumulative reward of each step is gathered along its subsequent steps
\begin{equation}
    R(s_k, a_k, i_k) = \sum_{i=k}^{K} \gamma^{i-k} r_k,
\end{equation}
\noindent where $K$ is the length of the trajectory, $\gamma$ is a discount factor.
With the cumulative aligned reward, although the lower steps are not able to get the structure perfectly matched, the subsequent steps can still get awarded by making correct decisions. Correspondingly, the lower steps will get a less severe penalty from the reward accumulation, which guarantees the flexibility in adjusting training signals.
Finally, by aligning the trajectory with the gold tree, we shall get a reward $R(s_k, a_k, i_k)$ for each step and a total reward for the trajectory:
\begin{equation}
     R(\tau) = \sum_{k=1}^{K}R(s_k, a_k, i_k).
\end{equation}

\paragraph{Optimization}

We aim to learn a stochastic policy of sentence selection module $\pi$ parameterized by $\theta$ which maximizes the expected cumulative reward:
\begin{equation}
    J_{\theta} = \mathbb{E}_{\tau \sim {\pi}_{\theta}(\tau)}[R(\tau)].
\end{equation}

Following \citet{Sutton1999PolicyGM}, we are able to approximate the gradient by sampling $N$ trajectories:

\begin{align}
 \nabla \bar J_{\theta} &= \mathbb{E}_{\tau \sim {\pi}_{\theta}(\tau)} [R(\tau) \nabla \log \pi_{\theta}(\tau)] \\
 &\approx \frac{1}{N} \sum_{n=1}^N \sum_{k=1}^{K_n} R(s_k^n, a_k^n, i_k^n) \nabla \log \pi_{\theta} (a_k^n \mid s_k^n),
\end{align}

\noindent where $K_n$ is the length of trajectory $\tau_n$, $s_k^n$ and $a_k^n$ denote the state and action at step $k$ in $\tau_n$.

\subsection{Deduction Generation}

For deduction generation, the model takes in two sentences and outputs a conclusion deduced over the input. 
We design two methods to generate deduction: model-based and rule-based. 
For model-based approach, we fine-tune a transformer-based \citep{Vaswani2017AttentionIA} Seq2Seq model to generate deduction on the fly during training and inference. Involving model-generated conclusions can provide more fluent and readable reasoning chains. 
However, unstable generation quality will introduce unexpected errors along the reasoning chains, which confuses the sentence selection module in the following steps.
For rule-based approach, no generation model is used in training.
We simply concatenate the chosen sentences with ``and'' to form the intermediate conclusions, which guarantees all information from the premises is well preserved. During inference, we additionally generate model-based deduction after the tree structure is fully predicted.
By default, we use the rule-based approach in training for the consideration of computational costs. We will elaborate on this more in Section \ref{abl-ded}.


\subsection{Overall Model Training}

Supervised learning before integrating RL in training can provide efficient parameter update with high-quality signals \citep{Silver2016MasteringTG}.
To help RL algorithm better converge on our task, we first apply supervised training on sentence selection with extracted gold sentence pairs of single steps from the EntailmentBank training set. 
During the training process of RL, we additionally apply scheduled sampling \citep{Bengio2015ScheduledSF} to alleviate the error propagation along the decision paths. 




\section{Experimental Settings}

\subsection{Dataset}
We evaluate our approach on EntailmentBank \citep{Dalvi2021ExplainingAW}, a benchmark that provides expert-annotated explanations of QA pairs in the form of entailment trees. Table~\ref{table_dataset} shows the detailed statistics of the dataset. EntailmentBank contains 1,840 questions along with 5,881 reasoning steps. 
QA pairs are randomly sampled from ARC dataset \citep{Clark2018ThinkYH} which consists of grade-school level science exam questions. The full fact corpus contains around 12K general-knowledge sentences derived from WorldTree V2 \citep{Xie2020WorldTreeVA}.

Depending on the composition of the given fact set $X$, the dataset offers three challenging settings. In \textbf{Task 1}, only gold facts are provided and will all serve as leaf nodes in the tree. In \textbf{Task 2}, for each QA pair, a total of 25 sentences are provided, including both gold facts and distractors. In \textbf{Task 3}, the most challenging setting, the model needs to first retrieve relevant facts from the full 
fact corpus, and then perform reasoning as in Task 2.

\begin{table}[]
\centering

\begin{tabular}{lcccc}
\hline
                & \textbf{Train} & \textbf{Dev} & \textbf{Test}  & \textbf{Total} \\ \hline
QA pairs        & 1,313 & 187 & 340   & 1,840 \\
Reasoning steps & 4,175 & 597 & 1,109 & 5,881 \\ \hline
\end{tabular}
\caption{\label{table_dataset}
Statistics of EntailmentBank dataset split.
}
\end{table}

\begin{table*}[!htbp]
\centering
\resizebox{\linewidth}{!}{
\begin{tabular}{clccccccc}
\hline
\multicolumn{1}{c}{\multirow{2}{*}{\textbf{Task}}} & \multicolumn{1}{c}{\multirow{2}{*}{\textbf{Method}}} & \multicolumn{2}{c}{\textbf{Leaves}} & \multicolumn{2}{c}{\textbf{Steps}} & \multicolumn{2}{c}{\textbf{Intermediates}} & \textbf{Overall} \\
\multicolumn{1}{c}{} & \multicolumn{1}{c}{}        & F1    & AllCorrect & F1   & AllCorrect & F1   & AllCorrect & AllCorrect \\ \hline
\multicolumn{1}{c}{\multirow{4}{*}{Task 1}} & EntailmentWriter (T5-11B)   & 99.0  & 89.4       & 51.5 & 38.2       & \textbf{71.2} & \textbf{38.5}      & \textbf{35.3}       \\
\multicolumn{1}{c}{} & EntailmentWriter (T5-Large) & 98.7  & 86.2       & 50.5 & 37.7       & 67.6 & 36.2       & 33.5       \\
\multicolumn{1}{c}{} & IRGR (T5-Large) & 97.6  & 89.4       & 50.2 & 36.8       & 62.1 & 31.8      & 32.4       \\
\multicolumn{1}{c}{}  & RLET (Our approach)  & \textbf{100.0} & \textbf{100.0}      & \textbf{54.6$_{0.4}$}   & \textbf{40.7$_{0.5}$}    & 66.9$_{0.3}$  & 36.3$_{0.4}$           &  34.8$_{0.3}$          \\ \hline

\multicolumn{1}{c}{\multirow{4}{*}{Task 2}} & EntailmentWriter (T5-11B)   & \textbf{89.1}  & \textbf{48.8}       & \textbf{41.4} & 27.7       & \textbf{66.2} & \textbf{31.5}      & 25.6       \\
\multicolumn{1}{c}{} & EntailmentWriter (T5-Large)  & 84.3  & 35.6       & 35.5 & 22.9       & 61.8 & 28.5       & 20.9       \\
\multicolumn{1}{c}{} & IRGR (T5-Large)   & 69.9  & 23.8      & 30.5 & 22.4       & 47.7 & 26.5       & 21.8       \\
\cline{2-9}
\multicolumn{1}{c}{}  & RLET (Our approach) 
& 81.0$_{0.9}$  & 39.0$_{1.4}$      & 38.5$_{0.3}$ & \textbf{28.4$_{0.3}$}       & 56.3$_{1.1}$  & 28.6$_{0.5}$ & \textbf{25.7$_{0.3}$}        \\ 
\hline

\multicolumn{1}{c}{\multirow{5}{*}{Task 3}}
& EntailmentWriter (T5-11B)   & \textbf{39.9}  & 3.8       & 7.4 & 2.9      & \textbf{35.9} & 7.1      & 2.9       \\
\multicolumn{1}{c}{} & EntailmentWriter (T5-Large) & 35.7  & 2.9       & 6.1 & 2.4       & 33.4 & 7.7       & 2.4       \\

\multicolumn{1}{c}{}  & RLET (Our approach)   & 38.3$_{0.3}$ & \textbf{9.1}$_{0.4}$   & \textbf{11.5}$_{0.4}$   & \textbf{7.1}$_{0.7}$  & 34.2$_{0.6}$ & \textbf{12.1}$_{0.6}$  & \textbf{6.9}$_{0.6}$    \\ 
\cline{2-9}
\multicolumn{1}{c}{} & IRGR (T5-Large) $^\diamond$  & 45.6  & 11.8 & 16.1 & \textbf{11.5} & 38.8 & \textbf{20.9}   & \textbf{11.5} \\
\multicolumn{1}{c}{}  & RLET (Our approach)$^\diamond$  & \textbf{47.2 }& \textbf{13.5}   &  \textbf{16.3}   & \textbf{11.5}  & \textbf{41.9} & 18.5     & \textbf{11.5}         \\ 

\hline

\end{tabular}
}
\caption{\label{table_main}
Experiment results on EntailmentBank test set. 
All baseline results come from published paper. $\diamond$ indicates using retrieval results produced by IRGR. We report average results and standard deviation across five runs.
}
\end{table*}

\subsection{Baselines}


\textbf{EntailmentWriter} \citep{Dalvi2021ExplainingAW} offers a strong baseline by linearising the tree structure and adopts sequence-to-sequence model to generate the whole tree along with intermediate conclusions in single pass. It has two versions, implemented on T5-11B (11 billion parameters) and T5-Large (770 million parameters) \citep{Raffel2020ExploringTL}. 

\noindent
\textbf{IRGR} \citep{Ribeiro2022EntailmentTE} designs an iterative retrieval-generation framework and improves the retrieval results on Task 3. For tree generation, it performs single step reasoning using T5-Large. 

\subsection{Implementation Details}

The sentence selection module is built with DeBERTa-v3-base model \citep{He2021DeBERTaV3ID}, which has 184M parameters in total. 
For Task 1, our model iteratively generates one-step reasoning until all given gold leaves are used. For Task 2 and Task 3, we add a special token \texttt{[END]} in action space at each step, and will stop the reasoning process once \texttt{[END]} token is selected. We further apply a fact filter trained on DeBERTa-v3-Large to help model identify relevant facts sentences in Task 2 and 3.
The default model-based deduction generation module is implemented with  BART-Large \citep{Lewis2020BARTDS} which has 406M parameters, trained with both ParaPattern \citep{Bostrom2021FlexibleGO} data and extracted gold premises along with intermediate conclusions.  ParaPattern contains synthetic data collected from Wikipedia, which has two sentences as input and a combined conclusion as output. For details on the ParaPattern data, we refer readers to their paper \citep{Bostrom2021FlexibleGO}. 
More hyperparameters can be found in Appendix~\ref{app_hyper}.

\subsection{Evaluation Metrics}

\citet{Dalvi2021ExplainingAW} introduces four standard metrics to automatically evaluate the entailment tree. As described in section~\ref{sec_reward}, the alignment algorithm is first applied to align the predicted tree $T_{pred}$ and the gold tree $T_{gold}$. We denote the aligned prediction tree as $T^{\prime}_{pred}$. 
In the following metrics, F1 measures the micro-average of the results and AllCorrect scores the whole tree as 1 in its corresponding metric if its F1 equals to 1. 


\paragraph{Leaves (F1, AllCorrect)}
evaluate how well the model performs in identifying facts that are relevant to questions and answers.
The F1 score is computed based on the selected leaf nodes in $T_{pred}$ and the gold leaf nodes. AllCorrect is 1 if they are perfectly matched, otherwise 0.

\paragraph{Steps (F1, AllCorrect)}
mainly evaluate the structure correctness of the trees. For each aligned step in $T^{\prime}_{pred}$, we measure whether its selected sentences (action in trajectory) matches the gold. F1 score is computed based on the number of perfectly matched steps.
We assign AllCorrect of 1 to a predicted tree if all steps in $T^{\prime}_{pred}$ exactly match with gold tree steps.

\paragraph{Intermediates (BLEURT, AllCorrect)} evaluate the generation quality of the intermediate conclusions. For each aligned step in $T^{\prime}_{pred}$, we define its intermediate conclusion $i$ is correct if $i$ has a BLEURT \citep{Sellam2020BLEURTLR} similarity score higher than 0.28 with the gold step conclusion. We assign AllCorrect of 1 if all intermediate conclusions in $T^{\prime}_{pred}$ are correct.

\paragraph{Overall (AllCorrect)}
From the above three metrics, we consider a predicted tree $T^{\prime}_{pred}$ as overall correct if the AllCorrect scores of its Leaves, Steps and Intermediates are all 1.



\begin{table*}[]
\centering 
\begin{tabular}{lcccccccc}
\hline
\multicolumn{1}{c}{\multirow{2}{*}{\textbf{Method}}}  &\multicolumn{2}{c}{\textbf{Leaves}} & \multicolumn{2}{c}{\textbf{Steps}} & \multicolumn{2}{c}{\textbf{Intermediates}} & \textbf{Overall} \\
\multicolumn{1}{c}{} & F1    & AllCorrect & F1   & AllCorrect & F1   & AllCorrect & AllCorrect \\ \hline

RLET  & \textbf{100.0} & \textbf{100.0}      & \textbf{54.6$_{0.4}$}   & \textbf{40.7$_{0.5}$}    & \textbf{66.9$_{0.3}$}  & \textbf{36.3$_{0.4}$}           &  \textbf{34.8$_{0.3}$} \\

~~~~w/o aligned reward & 100.0 & 100.0      & 53.9$_{0.2}$   & 40.2$_{0.1}$    & 66.7$_{0.4}$  &  36.2$_{0.2}$          &   34.6$_{0.2}$         \\ 
~~~~w/o RL & 100.0 & 100.0      & 50.1$_{0.4}$   & 37.8$_{0.2}$    & 64.4$_{1.0}$  &  34.6$_{0.4}$          &  33.0$_{0.4}$          \\ 
\hline
\end{tabular}

\caption{\label{table_abl}
Ablation results of Task 1 (no distractor) on EntailmentBank test set. 
}
\end{table*}

\section{Results \& Analysis}

\subsection{Main Results}



As shown in Table~\ref{table_main}, in Task 1 
RLET outperforms all baselines on the Steps metric, which verifies the strength of our design in improving structure correctness. Specifically, RLET yields an improvement of Steps F1/AllCorrect by absolute 3.1/2.5.
The backbone model that constructs the tree structure is DeBERTa-base with only 184M parameters, which is 60 times less than T5-11B. 
Our approach also achieves comparable results on intermediates and overall metrics with baselines. Note that in our default rule-based approach, our intermediate results come from plug-in deduction generation module, which can be improved with a stronger model and is out of the main scope of this work.

For Task 2 and Task 3, our framework outperforms all baselines on the most strict metric Overall AllCorrect, as shown in Table~\ref{table_main}. For fair comparison, we use the same retrieval results produced by \citet{Dalvi2021ExplainingAW} in Task 3.
Under the most challenging setting, RLET achieves significant improvement with 4.1/4.2 gain on Steps F1/AllCorrect, and outperforms all baselines on Overall AllCorrect with a score of 6.9.
With better retrieval results from IRGR \citep{Ribeiro2022EntailmentTE}, RLET still achieves higher Leaves performance.

\subsection{Ablation Study}

\paragraph{Sentence Selection}
To understand how each component contributes to our model, we conduct ablation studies on Task 1, as shown in Table~\ref{table_abl}.
We first analyze the influence of our crafted reward function with the alignment algorithm. By removing the cumulative aligned reward and assigning the reward solely based on exact match with gold trees, we witness a performance drop in Steps F1/AllCorrect. This indicates that our aligned reward function is effective in bridging the gap between training and evaluation. 
Furthermore, we test our approach without the entire RL training procedure, only keeping the supervised training model as the sentence selection module.
By training with extracted single steps, the performance decreases by a large margin (-4.5 in Steps-F1), which verifies the necessity of accumulating signals in training.


\paragraph{Deduction Generation}\label{abl-ded}

\begin{table*}[!htbp]
\centering
\resizebox{\linewidth}{!}{
\begin{tabular}{clccccccc}
\hline
\multicolumn{1}{c}{\multirow{2}{*}{\textbf{Task}}} & \multicolumn{1}{c}{\multirow{2}{*}{\textbf{Method}}} & \multicolumn{2}{c}{\textbf{Leaves}} & \multicolumn{2}{c}{\textbf{Steps}} & \multicolumn{2}{c}{\textbf{Intermediates}} & \textbf{Overall} \\
\multicolumn{1}{c}{} & \multicolumn{1}{c}{}        & F1    & AllCorrect & F1   & AllCorrect & F1   & AllCorrect & AllCorrect \\ \hline
\multicolumn{1}{c}{\multirow{3}{*}{Task 1}} & RLET  & 100.0 & 100.0      & 54.6$_{0.4}$   & \textbf{40.7$_{0.5}$}    & 66.9$_{0.3}$  & \textbf{36.3}$_{0.4}$           &  \textbf{34.8}$_{0.3}$          \\
\multicolumn{1}{c}{} & ~~ w/ T5 & 100.0  & 100.0 &   \textbf{54.8}$_{0.2}$   & 40.5$_{0.2}$ & \textbf{67.7}$_{0.1}$   & 36.1$_{0.3}$   &  34.0$_{0.3}$      \\
\multicolumn{1}{c}{} & ~~ w/ T5, w/o RL & 100.0  &  100.0  & 52.4$_{0.2}$  &  38.4$_{0.1}$ & 67.1$_{0.1}$ & 34.6$_{0.3}$  & 32.7$_{0.3}$    \\
 \hline

\multicolumn{1}{c}{\multirow{3}{*}{Task 2}} & RLET
& \textbf{81.0}$_{0.9}$  & \textbf{39.0}$_{1.4}$      & 38.5$_{0.3}$ & \textbf{28.4$_{0.3}$}       & 56.3$_{1.1}$  & 28.6$_{0.5}$ & \textbf{25.7$_{0.3}$}        \\ 
\multicolumn{1}{c}{} & ~~ w/ T5 & 80.4$_{0.4}$  & 37.7$_{0.4}$   & \textbf{39.3}$_{0.1}$ & 27.0$_{0.3}$   & \textbf{57.3}$_{0.3}$ & \textbf{29.0}$_{0.2}$   & 25.4$_{0.3}$       \\
\multicolumn{1}{c}{} & ~~ w/ T5, w/o RL   & 80.8$_{0.4}$ & 29.6$_{1.3}$      & 33.7$_{0.7}$ & 17.5$_{0.9}$      & 55.1$_{0.5}$  & 20.6$_{1.2}$       & 16.2$_{0.9}$        \\

\cline{2-9}
\hline

\multicolumn{1}{c}{\multirow{3}{*}{Task 3}}
& RLET  & 38.3$_{0.3}$ & \textbf{9.1}$_{0.4}$   & 11.5$_{0.4}$   & \textbf{7.1}$_{0.7}$  & \textbf{34.2}$_{0.6}$ & \textbf{12.1}$_{0.6}$  & \textbf{6.9}$_{0.6}$     \\
\multicolumn{1}{c}{} & ~~ w/ T5 & \textbf{38.7}$_{0.4}$ & 7.8$_{0.4}$ & \textbf{11.9}$_{0.3}$ & 6.5$_{0.2}$ & 32.7$_{0.4}$  & 10.4$_{0.4}$ & 6.5$_{0.2}$     \\
\multicolumn{1}{c}{} & ~~ w/ T5, w/o RL & 36.6$_{0.3}$ &  3.7$_{0.3}$  & 8.8$_{0.3}$ & 3.4$_{0.2}$  & 33.0$_{0.1}$ & 6.9$_{0.6}$   & 3.4$_{0.2}$      \\

\hline

\end{tabular}
}
\caption{\label{table_deduction_abl}
Ablation results on EntailmentBank test set with model-generated deduction in the training process. ``RLET'' denotes using rule-based deduction in training,
``w/ T5'' means using T5-deduction during training, 
``w/ T5, w/o RL'' means using T5-deduction during inference but without RL training on the sentence selection module.
}
\end{table*}


The deduction generation module plays an important role in ensuring the fluency and readability of the reasoning chains. We experiment with model-based and rule-based approaches and find that the latter yields better results.
In training, we observe that the finetuned BART model is likely to repeat one of the input premises thus losing useful information \citep{Bostrom2021FlexibleGO}, which hampers the convergence of the sentence selection module in training.

To take a step further, we involve a stronger deduction generation module from MetGen \citep{Hong2022METGENAM} into the training process, which is trained on T5-Large \citep{Raffel2020ExploringTL} with 770M parameters.
The reasoning module of MetGen is trained with additional synthetic data and requires the specific reasoning pattern as a prompt in its input. To get such a pattern for each reasoning step, we train a pattern selector with DeBERTa using the manually annotated labels to specify the sentence pair as one of substitution, conjunction or if-then patterns. 
Improving the generation quality is orthogonal to our work, so we reuse the released weight of MetGen's reasoning module.\footnote{\url{https://github.com/Raising-hrx/MetGen}}  We call this setting as ``T5-deduction''. More details can be found in Appendix \ref{app-ded}.
Experiment results in Table~\ref{table_deduction_abl} show that simple rule-based models can achieve comparable performance with model-based  ``T5-deduction''. Including model-generated conclusions in training process can offer more fluent textual reasoning chains, thus improving Intermediates scores. However, a large and comprehensive deduction generation module is needed to effectively retain the information in the form of natural language without sacrificing overall performance.

\subsection{In-hoc Reasoning}


The original settings of EntailmentBank contain hypothesis as the guidance of post-hoc explanation generation.
However, in most practical scenarios where only questions are available, the model should ideally reason over knowledge facts to derive the answer while generating an explainable reasoning path. This is defined as Open-ended Commonsense Reasoning (OpenCSR) in \citet{Lin2021DifferentiableOC}.
With this intuition, we therefore investigate the potential of entailment trees in generating explanations by substituting hypothesis with questions in sentence selection. The model needs to additionally generate 
the root node as the final conclusion, instead of copying the hypothesis.
We conduct in-hoc reasoning experiments on Task 1 and Task 2, simplifying the OpenCSR task by narrowing down the available knowledge facts.

We evaluate the structure correctness of the generated explanations in Table~\ref{table_nohypo}.
We observe that baseline models (without RL) encounter a performance drop comparing to hypothesis-guided post-hoc setting.
It suggests that the answers do offer a strong guidance in generating explanations. 
With RL, RLET achieves Steps AllCorrect of 38.5, which performs on par with post-hoc Entailment-Writer results. 
Greater drop in Task 2 indicates that the removal of hypothesis confounds the model in identifying relevant sentences, resulting in a poor performance across the board. 
More details can be found in Appendix~\ref{app_ques}.
Our initial observations address the difficulty of question-guided in-hoc reasoning. 
A comprehensive OpenCSR system should further include the retrieval of relevant sentences based on questions and proper answer evaluation. 
We leave these for future work.

\begin{table*}[]
\centering
\begin{tabular}{clccccccc}
\hline
\multicolumn{1}{c}{\multirow{2}{*}{\textbf{Task}}} & \multicolumn{1}{c}{\multirow{2}{*}{\textbf{Method}}} & \multicolumn{2}{c}{\textbf{Leaves}} & \multicolumn{2}{c}{\textbf{Steps}} & \multicolumn{2}{c}{\textbf{Intermediates}} & \textbf{Overall} \\
\multicolumn{1}{c}{}        & \multicolumn{1}{c}{}        & F1    & AllCorrect & F1   & AllCorrect & F1   & AllCorrect & AllCorrect \\ \hline
\multicolumn{1}{c}{\multirow{2}{*}{Task 1}} & RLET    & 100.0  & 100.0  & 50.1 & 38.5       & 46.7 & 23.5       & 22.9       \\
\multicolumn{1}{c}{}        &~~~~w/o RL   & 100.0  & 100.0       & 46.9 & 34.4       & 44.2 & 22.7      & 20.6       \\
\hline
\multicolumn{1}{c}{\multirow{2}{*}{Task 2}} & RLET    & 68.1  & 20.9  & 29.4 & 17.7 & 32.5 & 14.7    & 11.8       \\
\multicolumn{1}{c}{}        &~~~~w/o RL   & 70.4  & 18.5  & 25.0 & 11.5  & 32.4 & 10.6 & 7.7    \\
\hline
\end{tabular}
\caption{\label{table_nohypo}
Experiment results on EntailmentBank test set under in-hoc reasoning setting.
}
\end{table*}

\subsection{Data Efficiency}

In practice, high-quality explanation annotations can be costly to obtain, which makes it difficult to train large scale models.
An ideal system is expected to have great generalisability even with few annotated explanations for training.
To evaluate how RLET can benefit from the RL algorithm under this setting, we experiment with less data under Task 1. We divide the data based on the number of given facts for each QA pair.
Detailed statistics of the data split are shown in Table~\ref{table_scarce}. 

\begin{figure}[]
    \centering
    \hspace*{-25pt}%
    \includegraphics[width=1.25\linewidth]{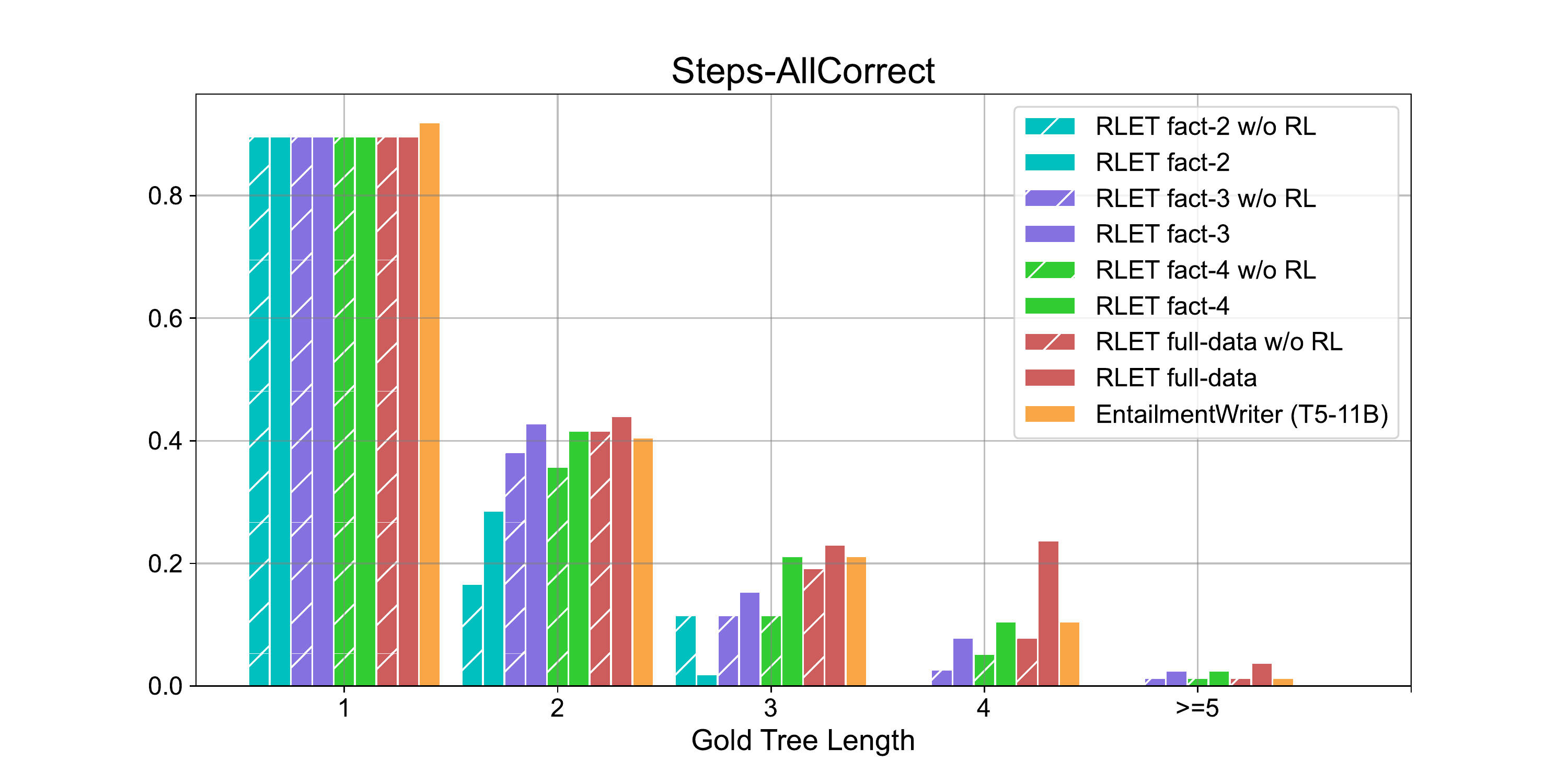}
    \vspace{-5pt}
    \caption{\label{fig_len} 
    Results broken down by the length of the gold trees on the test set in Task 1.
    Bars in the same color represent the same data used for training. Bars with shading indicate the removal of RL.
    }
\end{figure}

We break down the results by the length of the gold trees in Figure~\ref{fig_len}.
Results show that with only 35.6\% of the total training steps, RLET (fact-4) achieves comparable performance with the full-data trained baselines on Steps metric, indicating our approach can help model generalise with fewer data.
It is also indicated that RL framework is able to offer more gains as the length increases.

\begin{table}[]
\centering
\resizebox{\linewidth}{!}{
\begin{tabular}{cccc}
\hline
\textbf{Setting }& \textbf{\# Facts}         & \textbf{QA pair}       & \textbf{Reasoning step} \\
\hline
fact-2  & $\leq$2 & 20.6\%   & 6.5\%   \\
fact-3  & $\leq$3 & 43.9\%   & 20.0\%  \\
fact-4  & $\leq$4 & 62.3\%   & 35.6\%  \\
full data    & full data     & 100\%  & 100\%  \\
\hline
\end{tabular}
}
\caption{\label{table_scarce}
Details of data efficiency settings. 
}
\end{table}

\section{Related Work}

\paragraph{Tree Structured QA Explanations}


Existing methods on generating entailment trees for QA explanations can be categorized into two branches: single pass generation and stepwise generation. 
Single pass methods \citep{Dalvi2021ExplainingAW} model the tree structure as a linearised chained sequence to fit in the Seq2Seq model. Alternatively, more recent work tends to generate step-by-step reasoning, performing fine-grained deduction iteratively. 
Among them, one line of stepwise methods jointly select steps and generate intermediate conclusions. IRGR \citep{Ribeiro2022EntailmentTE} puts effort in designing an iterative retrieval-generation approach, where one step reasoning is generated at a time through a Seq2Seq model.
Another line of stepwise methods handle sentence selection and intermediate conclusion generation with separate modules, which is closely related to our approach.
SCSearch \citep{Bostrom2022NaturalLD} models the tree construction as a best-first search using heuristic guidance. However, their approach is mainly evaluated on the accordance between final conclusions and the hypothesis, which varies from standard evaluation metrics.
Another concurrent work MetGen \citep{Hong2022METGENAM} also separates these two parts and trains the generation module with additional human annotations and more synthetic data from Wikipedia. While all previous works focus on forward reasoning, MetGen further shows the effect of backward abductive reasoning. 
Existing methods are all trained with individual signals from extracted gold steps.
In that light, RLET is complementary, as it models the reasoning process as an MDP in which the stepwise modules can be a component.


The deduction generation module of RLET can be connected with proof generation work that generates natural language logical deductions from premises or rules \citep{Saha2020PRoverPG, Tafjord2021ProofWriterGI, Sanyal2022FaiRRFA}. To improve the generation quality, \citet{Bostrom2022NaturalLD} propose ParaPattern by creating synthetic logical deduction data from Wikipedia, which serves as a good supplement to train the generative model.

Different from existing approaches, RLET models the cumulative signals across the whole tree in training, benefiting from our designed reward function under reinforcement learning framework.

\paragraph{Path Reasoning using Reinforcement Learning}
Our work also aligns well with multiple automated reasoning tasks built with RL \citep{Xian2019ReinforcementKG, Liu2021ImprovingPM,  Poesia2021ContrastiveRL}.
Especially, reinforcement learning has exhibited its attractiveness in knowledge graph reasoning \citep{Xiong2017DeepPathAR, Das2018GoFA, Lin2018MultiHopKG}, where the multi-hop path can be represented as sequential decision problems. 
Similar to the above RL methods, we formulate the entailment tree generation task as a trajectory of reasoning steps. In contrast to KG based multi-hop reasoning path, however, RLET generates intermediate conclusions of the reasoning steps, providing fluent natural language explanations in detail.




\section{Conclusion}

We presented RLET, a RL-based entailment tree generation framework, which contains sentences selection and deduction generation modules and 
can be trained with cumulative signals across the entire reasoning tree. Experiments show that RLET outperforms existing baselines on structure correctness and is applicable in practical scenarios. 
Future directions include applying RL framework on other stepwise methods with more stable and sophisticated RL algorithms.

\section*{Limitations}

First, sentences are likely to be used more than once when reasoning in real practice. RLET removes used sentences at each time step to reduce the size of action space, which leads to a performance loss of 9.4\% on the overall All-Correct. 
Second, in the sentence selection module, RLET always picks two sentences to merge, while the original dataset contains multi-sentence steps. Though this harms the evaluation results as discussed in Appendix \ref{sec:appendix}, this is a minor limitation because the reasoning format is not strictly standardized in real practice.
Furthermore, adding \texttt{[END]} token to action space and applying additional fact filter in distractors settings  is a naive approach and leaves room for further improvement.
Finally, as vanilla policy gradient method is sensitive to hyperparameters and can have large variance, we leave the exploration of more stable RL algorithms in reasoning for future work.



\section*{Ethics Statement}
Explainable QA is an important branch in the field of Question Answering.
The data used in our work all comes from public dataset, and our proposed method can be further integrated into other stepwise systems.
Our work is conformant to ACL Code of Ethics.

\section*{Acknowledgements}
We would like to thank the anonymous reviewers for their valuable suggestions and feedback. This work was supported by the National Key Research and Development Program of China (No.2020AAA0106700) and National Natural Science Foundation of China (No.62022027). 

\bibliography{anthology,custom}
\bibliographystyle{acl_natbib}

\appendix

\section{Hyperparameters} \label{app_hyper}

For pre-RL supervised training, we set a learning rate of 2e-5, a batch size of 2 and train the model for 20 epochs. 
For RL training, we set the discount factor $\gamma$ as \{0.9, 0.99\}, initial learning rate as 1e-5, warmup ratio as 0.05, and train the model for 20 epochs. $\gamma = 0.99$ yields better results. The scheduled sampling ratio decays linearly from 1.0 to 0.5. The total RL training costs 6 hours.

The fact filter in Task 2 and 3 is trained with an initial learning rate of 1e-5, warm up ratio of 0.1, for 10 epochs. In Task 2 we save top 5 sentences and filter out sentences with similarity scores lower than 0.98 in Task 3, which are selected based on the validation set.

The deduction generation module is implemented with BART-Large \citep{Lewis2020BARTDS} which has 406M parameters. We first train it on the ParaPattern data with an initial learning rate of 3e-5, batch size of 16 for one epoch, and further finetune the trained model on extracted gold steps for another two epochs with an initial learning rate of 3e-5, batch size of 16. The learning rate decays with linear scheduler. The total training costs 2 hours.

We use AdamW \citep{Loshchilov2019DecoupledWD} for optimization, with $\beta_1=0.9, \beta_2=0.99$. All experiments are conducted on 2 Tesla T4 GPUs with 16GB memory. 

\section{Manual Evaluation}
\label{sec:appendix}

The automatic evaluation is an underestimation of our approach because RLET only selects two sentences per action while 25.88\% trees in test set contain multiple-premise (more than 2 premises) reasoning steps, which will result in 0 in Steps/Overall metrics. To provide a more flexible comparison, we manually annotate the first 50 trees in Task 1 test set on structure validity in Table~\ref{table_manual}. The automatic evaluation results are the Steps-AllCorrect scores, while we manually consider a tree is valid in structure if the predicted tree can also perform valid reasoning under its structure. The results indicate RLET can outperform the strongest baseline under both automatic and manual evaluation.

\begin{table}[]
\resizebox{\columnwidth}{!}{
\begin{tabular}{lcc}
\hline
Method & Automatic  & Manual  \\
\hline
RLET      & 0.48   & 0.76   \\
EntailmentWriter (T5-11B) & 0.44   & 0.70   \\
\hline
\end{tabular}
}
\caption{\label{table_manual}
Automatic and manual evaluation results on structure validity of the first 50 trees in Task 1 test set.
}
\end{table}

\section{Evaluation Details in Question-guided Reasoning}\label{app_ques}
In Table~\ref{table_nohypo}, we witness a large performance drop comparing to hypothesis-guided reasoning, especially on Intermediate and Overall metrics. 
In hypothesis-guided reasoning, the generated reasoning chain always ends with "hypothesis". Consequently, the last step that contains "hypothesis" as intermediate conclusion is always labeled as "correct" under Intermediates metric. In contrast, when we remove hypothesis, the model has to generate the last intermediate conclusion instead of directly copying the hypothesis. 
Hence, the model suffers from the errors accumulated by the deduction generation and the BLEURT evaluation method, and struggles to draw a correct final conclusion that will be compared with gold hypothesis.
Cumbersome as it is, our main idea lies in the structure correctness under question-guided reasoning, which can be fulfilled with Leaves and Steps metrics. How to measure the answer correctness in the reasoning chain remains to be further explored in future work.





\section{Details of T5-deduction Setting}\label{app-ded}

Authors of MetGen manually annotated the reasoning patterns of 400 separate steps in the training set, and 275 steps in the validation set. 
We adopt these annotations to finetune a DeBERTa-Large model as our pattern selector, which takes in two premise sentences as input and predicts its corresponding reasoning pattern as substitution, conjunction or if-then. The pattern selector achieves an accuracy of 81.5\% on annotated sentences in the validation set. 
In Table~\ref{table_deduction}, we evaluate the generation quality of different deduction generation modules. Given a pair of premise sentences, if the generated conclusion has a BLEURT score greater than 0.28, then it is considered as 1 on ACC.
During training, we save the deduction generation results into a cache to accelerate the training process. The parameters of both pattern selector and MetGen's reasoning module are fixed during the training of the sentence selection module.

\begin{table}[]
\resizebox{\columnwidth}{!}{
\begin{tabular}{lcc}
\cline{1-3}
                              & ACC  & BLEURT \\ \hline
RLET BART-Large              & 77.3 & 59.5                 \\
MetGen (w/ predicted pattern) & 80.7 & 62.2                 \\
MetGen (w/ gold pattern)      & 84.3 & 65.5                 \\ \hline
\end{tabular}
}
\caption{\label{table_deduction}
Generation quality of deduction generation modules on the manually annotated sentences from validation set. BLEURT denotes the average BLEURT scores across all data.
}
\end{table}

\section{Case Study}

In this section, we illustrate some examples from Task 1 test set in which the predicted tree can perform a valid reasoning in a different structure with the gold tree. 

\begin{figure}[]
    \centering
    \includegraphics[width=1.0\linewidth]{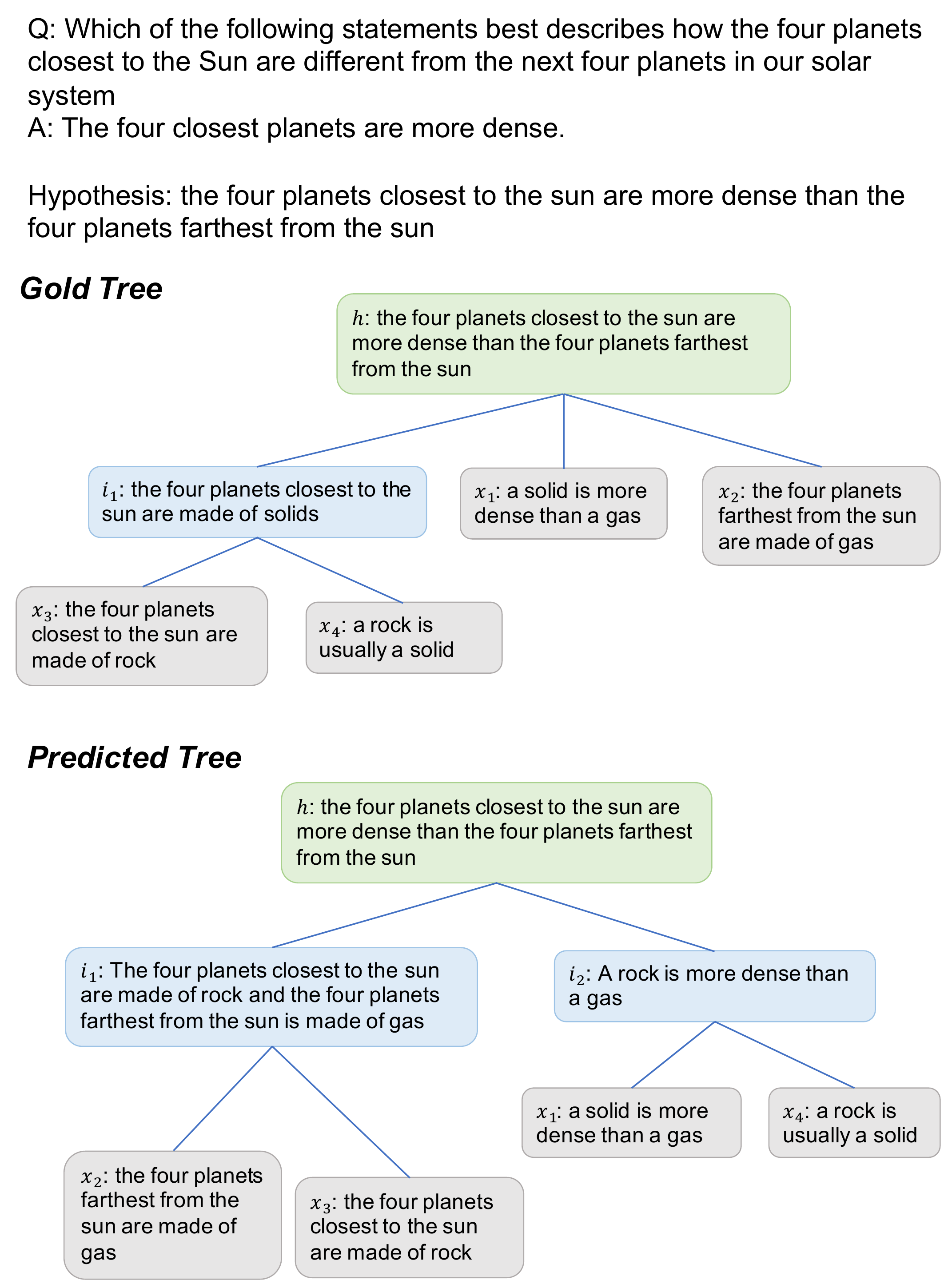}
    \vspace{-10pt}
    \caption{\label{fig_case1} An example of misevaluation in which the gold tree contains a three-premise reasoning step. Though not exactly matched with gold, the predicted tree can also fulfill the reasoning process with two-premise steps.}
\end{figure}


\begin{figure}[]
    \centering
    \includegraphics[width=1.0\linewidth]{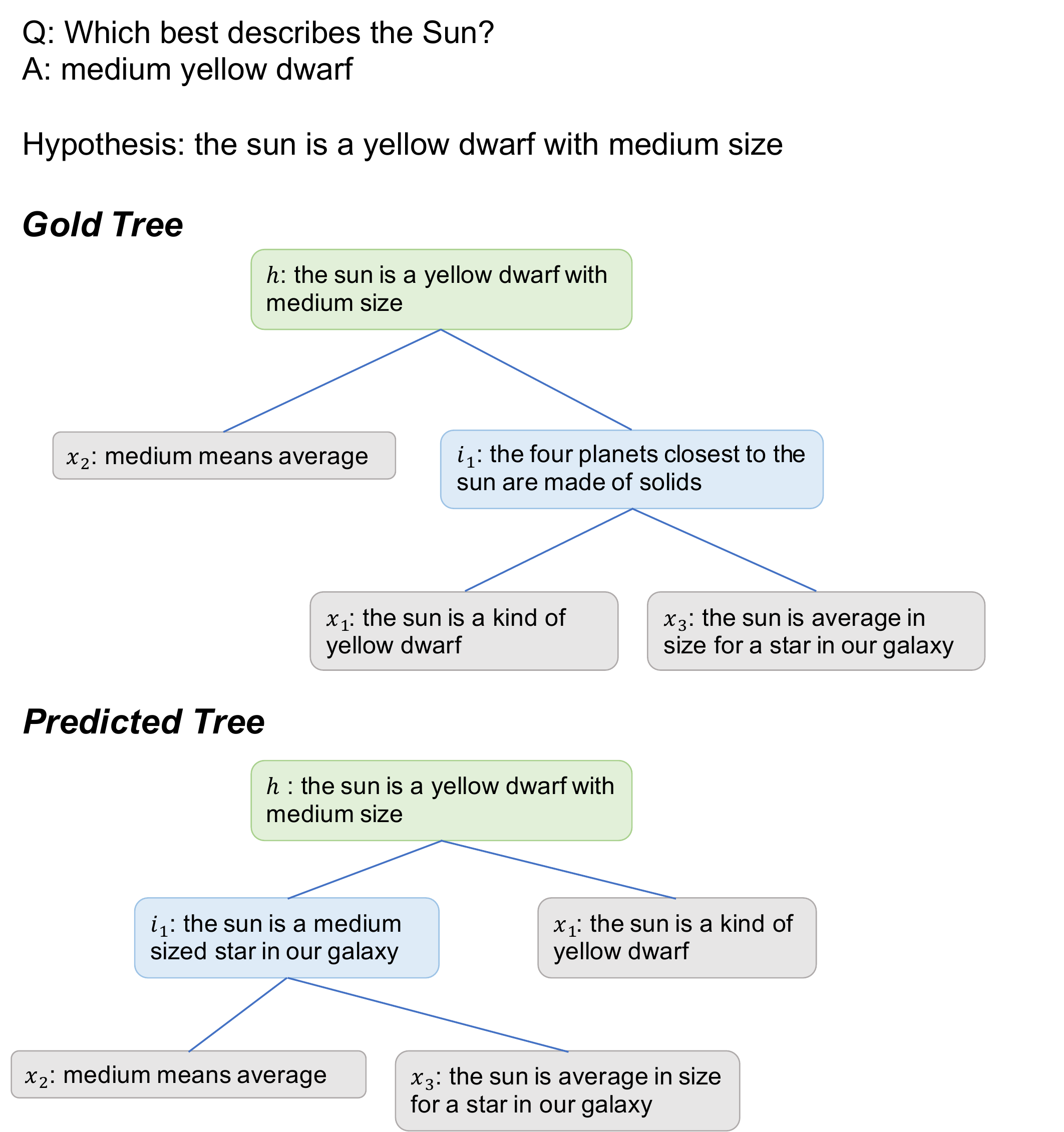}
    \vspace{-10pt}
    \caption{\label{fig_case3} An example of misevaluation in which gold tree and predicted tree vary in the order of the sentence composition. Both can provide valid explanations in the form of binary entailment trees.}
\end{figure}

\end{document}